\documentclass[final,5p,times,authoryear]{elsarticle}

\usepackage{amssymb}
\usepackage{amsmath}
\usepackage{amsthm}
\usepackage{bm}
\usepackage{hyperref}
\usepackage{booktabs}
\usepackage{multirow}
\usepackage{subcaption}
\usepackage{arydshln}
\usepackage{graphicx}
\usepackage[table,xcdraw]{xcolor}
\usepackage[normalem]{ulem}
\useunder{\uline}{\ul}{}
\usepackage[ruled,vlined]{algorithm2e}

\theoremstyle{definition}
\newtheorem{definition}{Definition}[section]

\journal{}

\begin{document}

\begin{frontmatter}



\title{Robust Learning on Heterogeneous Graphs with Heterophily: A Graph Structure Learning Approach} 


\author[1]{Yihan Zhang} 
\ead{zyh23@mails.tsinghua.edu.cn}

\author[1]{Ercan E. Kuruoglu\corref{cor1}} 
\ead{kuruoglu@sz.tsinghua.edu.cn}
\cortext[cor1]{Corresponding author.}

\affiliation[1]{organization={Shenzhen International Graduate School, Tsinghua University},
            city={Shenzhen},
            country={China}}

\begin{abstract}
Heterogeneous graphs with heterophily have emerged as a powerful abstraction for modeling complex real-world systems, where nodes of different types and labels interact in diverse and often non-homophilous ways. Despite recent advances, robust representation learning for such graphs remains largely unexplored, particularly in the presence of noisy or misleading connectivity.
In this work, we investigate this problem and identify structural noise as a critical challenge that significantly degrades model performance.
To address this issue, we propose a unified framework, Heterogeneous Graph Unified Learning (HGUL), which jointly handles heterophily and noisy graph structures.
The framework consists of three complementary modules: a kNN-based graph construction module that recovers reliable local neighborhoods, a graph structure learning module that adaptively refines the adjacency by filtering noisy edges, and a heterogeneous affinity learning module that captures class-level relationships via an extended affinity matrix derived from a polynomial graph kernel.
Extensive experiments on multiple datasets demonstrate that HGUL consistently outperforms existing methods on clean graphs and maintains strong robustness under varying levels of structural noise. The results further underscore the importance of jointly modeling heterophily and noise in heterogeneous graph learning.
\end{abstract}




\end{frontmatter}



\section{Introduction}
Heterogeneous graph neural networks (HGNNs) have demonstrated strong capability in learning from heterogeneous graphs and have been widely studied in prior work \citep{dong2017metapath2vec,wang2019heterogeneous,zhang2019heterogeneous,hu2020heterogeneous}. Early models and frameworks largely inherit the homophily assumption from homogeneous graph learning, where connected nodes are expected to share similar labels or features.
However, recent studies have challenged this assumption. Prior work \citep{zhu2020beyond,zhu2021graph} shows that, in graph neural networks, nodes with different labels may still contribute to each other in varying degrees. Building on this insight, more recent research investigates heterophily in heterogeneous graphs and demonstrates both theoretically and empirically that it widely exists in real-world scenarios \citep{lin2025heterophily,li2025heterophily,shen2025heterophily}.
For example, in academic networks, papers from different research areas may frequently cite each other, especially in interdisciplinary fields, forming heterophilous connections. In recommendation systems, users may interact with items from diverse categories that do not share similar attributes.
These examples illustrate that connected nodes in heterogeneous graphs may exhibit complementary rather than similar semantics, which challenges conventional graph learning paradigms.

\begin{figure}[t]
\centering
\includegraphics[width=0.9\linewidth]{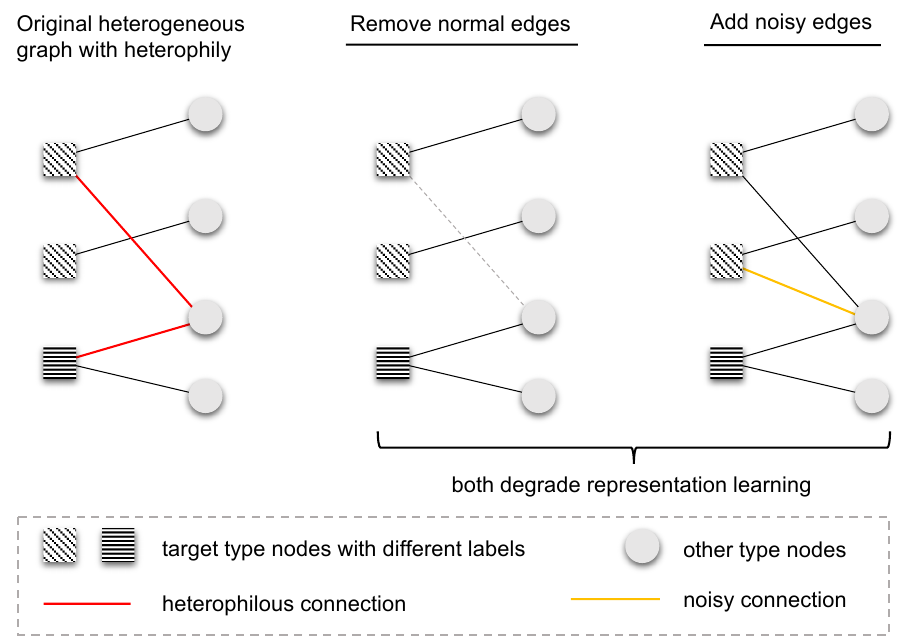}
\caption{Illustration of the studied problem}
\label{fig:illus-h2}
\end{figure}
In addition to heterophily, heterogeneous graphs also require careful consideration of noisy connectivity and robustness. Prior studies have shown that real-world heterogeneous graphs often contain misleading or noisy edges, which can propagate incorrect information across node types and relations \citep{zhang2022robust}. 
Despite these challenges, robust learning specifically tailored for heterogeneous graphs with heterophily remains largely underexplored. We will discuss and provide solutions under this setting.

\begin{figure}
\centering
\begin{subfigure}[]{0.45\linewidth}
    \centering
    \includegraphics[width=\linewidth]{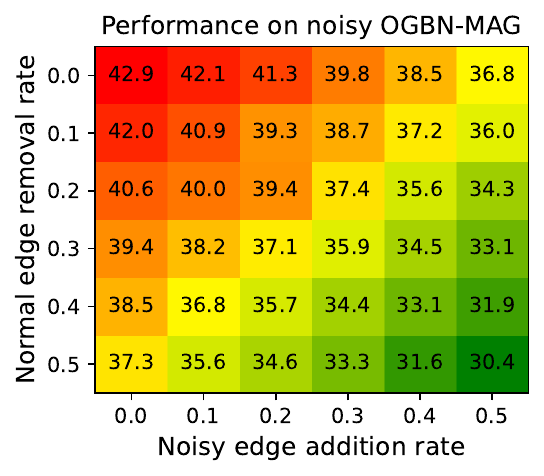}
    \caption{GCN}
\end{subfigure}
\begin{subfigure}[]{0.45\linewidth}
    \centering
    \includegraphics[width=\linewidth]{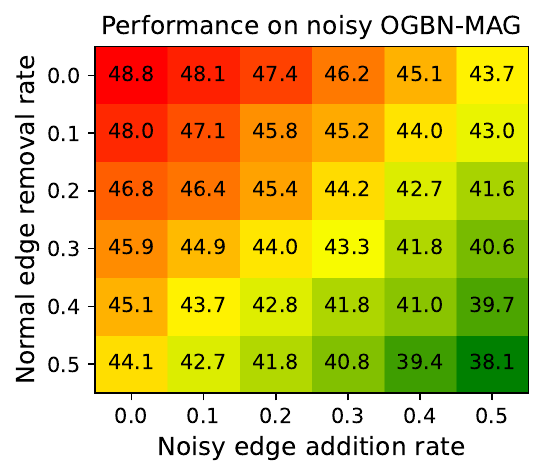}
    \caption{R-GCN}
\end{subfigure}
    \caption{Performance of different models w.r.t. removing positive and negative edges on ogbn-mag.}
    \label{fig:robust-het}
\end{figure}

\paragraph{Empirical findings}
Intuitively, perturbations to the graph structure—either by adding or removing edges—can significantly affect representation learning quality (Figure \ref{fig:illus-h2}).
To highlight the necessity of robust learning in this setting, we conduct an empirical analysis. 
Specifically, we perform controlled experiments on the ogbn-mag dataset \citep{hu2020open}, a representative heterogeneous graph exhibiting substantial heterophily across multiple node and relation types. We simulate noisy connectivity by introducing both edge removal and edge addition under the original meta-relation schema, ensuring that added edges remain type-consistent while introducing structural perturbations.
By varying the perturbation rate from 0 to 0.5, we generate graphs with progressively increasing noise levels. We then evaluate two representative models, GCN and R-GCN, under these conditions, and report the results in Figure~\ref{fig:robust-het}.
The results show that learning on heterogeneous graphs with heterophily is highly sensitive to noisy connectivity. As the noise level increases, model performance degrades significantly, indicating that existing methods lack robustness to structural perturbations in this setting. Moreover, due to the complexity of heterogeneous structures, existing approaches cannot be directly extended from homogeneous settings. These observations suggest that robust learning for heterogeneous graphs with heterophily requires dedicated design.

To address this gap, we propose an interpretable framework, \underline{H}eterogeneous \underline{G}raph \underline{U}nified \underline{L}earning (HGUL), for robust representation learning on heterogeneous graphs with heterophily. The proposed framework consists of three complementary components:
A kNN-based learning module, which constructs a similarity-driven graph to ensure robust feature aggregation when the original edge set is unreliable, while also recovering potentially beneficial connections;
A graph structure learning module, which assigns learnable weights to edges and filters out noisy connections through adaptive sparsification;
A heterogeneous affinity learning module, which captures class-level relationships by constructing a heterogeneous affinity matrix, enabling effective modeling of heterophilous interactions.
By integrating these components, HGUL jointly addresses both structural noise and heterophily, leading to more robust and expressive representations.

Our contributions are concluded as follows:
\begin{itemize}
    \item We consider robust learning on heterogeneous graphs with heterophily for the first time and propose a lightweight and scalable framework, HGUL. The kNN construction and structure learning modules effectively remove noisy edges while recovering informative connections.
    \item We introduce a heterogeneous affinity learning module that explicitly models class-level relationships. Combined with the other components, HGUL jointly mitigates the negative effects of noise and heterophilous connections.
\item Extensive experiments demonstrate that HGUL consistently improves prediction performance and robustness under various noise levels, while maintaining manageable computational cost and scalability.
\end{itemize}

\section{Related Works}
\subsection{Robust Graph Learning}
Prior to graph structure learning (GSL), noisy edges were typically handled via graph sparsification. For example, \citep{jiang2021pre} prunes edges in large-scale heterogeneous graphs based on personalized PageRank (PPR) scores. While effective in reducing computational cost and noise, such approaches operate purely on structural heuristics and remain agnostic to downstream learning objectives, often leading to suboptimal graphs for GNN-based tasks.
GSL addresses this limitation by treating the adjacency matrix as a latent and learnable component rather than a fixed input. A dominant approach adopts continuous, differentiable parameterizations, modeling edges as real-valued affinities jointly optimized with GNN parameters \citep{franceschi2019learning, chen2020iterative}. These methods often alternate between structure refinement and representation learning to mitigate error accumulation. To prevent overfitting and enforce desirable properties, another line of work introduces explicit structural priors, such as spectral regularization, to encourage smoothness and sparsity \citep{kumar2019structured}.
More recently, GSL has been extended to heterogeneous graphs, where both node and relation types must be considered. A representative work, HGSL \citep{zhao2021heterogeneous}, introduces a unified framework that jointly learns heterogeneous graph structures and node embeddings. Specifically, it designs feature-level and semantic-level graph generators to capture both local similarity and high-level relational semantics.


\subsection{Learning on Heterogeneous Graph with Heterophily}
Homophily refers to the tendency of connected nodes to share similar features or labels, under which neighborhood aggregation in GNNs is highly effective. In contrast, heterophily describes graphs where linked nodes often exhibit dissimilar features or labels \citep{zhu2020beyond, pei2020geom, lim2021new, lim2021large}, posing significant challenges to standard message-passing schemes.
Extending heterophily to heterogeneous graphs introduces additional complexity. In such settings, dissimilarity arises not only from label mismatch but also from diverse node and edge types, making the interaction patterns more intricate. When heterogeneity and heterophily coexist, traditional HGNNs, which often assume consistent semantic alignment across relations, can become ineffective.
Recent work highlights that heterophily is prevalent in real-world heterogeneous graphs and varies across relation types \citep{lin2025heterophily}. Different relations may exhibit distinct homophily levels, motivating more fine-grained modeling. Meta-path–based analysis has emerged as a key tool in this context, as it captures typed relational semantics and enables the measurement of homophily or heterophily along specific relational patterns rather than assuming a global property.
Building on this insight, existing methods can be broadly categorized into structure-level and model-level approaches. Structure-level methods aim to reshape the graph to better suit homophily-based learning. For example, HDHGR \citep{guo2023homophily} introduces meta-path–induced homophily metrics to characterize heterogeneous graphs and performs relation-aware graph rewiring, selectively adding or removing edges to enhance homophilic structures.
In contrast, model-level approaches directly design heterophily-aware architectures without modifying the graph. Hetero2Net \citep{li2025heterophily} proposes a framework that leverages auxiliary self-supervised tasks, such as masked meta-path prediction and masked label prediction, to improve representation learning under heterophily. Other recent works \citep{shen2025heterophily, lin2025heterophily} further explore disentangled representations, adaptive aggregation, or Transformer-based designs to capture both homophilic and heterophilic signals across relations.

Overall, these studies suggest that learning on heterogeneous graphs with heterophily requires moving beyond uniform aggregation, toward relation-aware, pattern-specific modeling and robustness to mixed structural signals.

\subsection{Combining Noise with Heterophily}
Traditional robust graph learning methods primarily focus on denoising or refining graph structures under the homophily assumption, which can fail under heterophily.
Recent work begins to bridge this gap by explicitly modeling robustness in heterophilic settings. For example, SparseGAD \citep{gong2023beyond} jointly learns structure and representations while accounting for both homophilic and heterophilic connections.
\citep{wu2024robust} involves a trainable edge discriminator that divides the original graph into homophilic and heterophilic views, followed by dual-channel encoders (with low- and high-pass filters) and unsupervised contrastive learning to derive robust node representations.
More recently, \citep{xie2025robust} leverages high-pass graph filtering to emphasize feature differences between neighbors, followed by an adaptive norm-based graph learning procedure to handle varying noise levels.
These works are closely related to theoretical findings that connect heterophily with robustness. Prior studies show that heterophily fundamentally alters the effect of structural perturbations and that architectures designed for heterophily (e.g., separating ego and neighbor embeddings \citep{zhu2020beyond}) can also enhance robustness against adversarial attacks \citep{zhu2022does}.
Overall, combining robustness with heterophily introduces a key challenge: distinguishing harmful noise from informative heterophilic edges.
Despite recent progress, current methods are still limited in elaborating the essential differences between two factors and lack research on heterogeneous graphs.

\section{Methodology}

\subsection{Preliminaries}
Unlike homogeneous graphs, heterogeneous graphs contain multiple types of nodes or relations, enabling the modeling of complex, multi-relational data.
A heterogeneous graph (or heterogeneous information network) is defined as $\mathcal{G} = (\mathcal{V}, \mathcal{E}, \phi, \psi)$, where $\mathcal{V}$ and $\mathcal{E}$ denote the sets of nodes and edges, respectively. The functions $\phi: \mathcal{V} \rightarrow \mathcal{A}$ and $\psi: \mathcal{E} \rightarrow \mathcal{R}$ map each node and edge to its corresponding type, with $\mathcal{A}$ and $\mathcal{R}$ representing the sets of node and relation types.
Each node $v \in \mathcal{V}$ is associated with a feature vector $\mathbf{x}*v \in \mathbb{R}^{d*{\phi(v)}}$, where the feature dimension may vary across node types. The features of nodes of the same type constitute a feature matrix $X_t$.

In this work, we focus on node classification in heterogeneous graphs. Given a target node type $t$, the goal is to learn a function $f: \mathcal{V}_t \rightarrow \mathcal{Y}$ that predicts labels for nodes of type $t$. During training, labels are available only for a subset of nodes $\mathcal{V}_L \subset \mathcal{V}_t$, and the objective is to generalize to unlabeled nodes $\mathcal{V}_U = \mathcal{V}_t \setminus \mathcal{V}_L$.



\subsection{Overall Learning Framework}
\begin{figure}
    \centering
    \includegraphics[width=\linewidth]{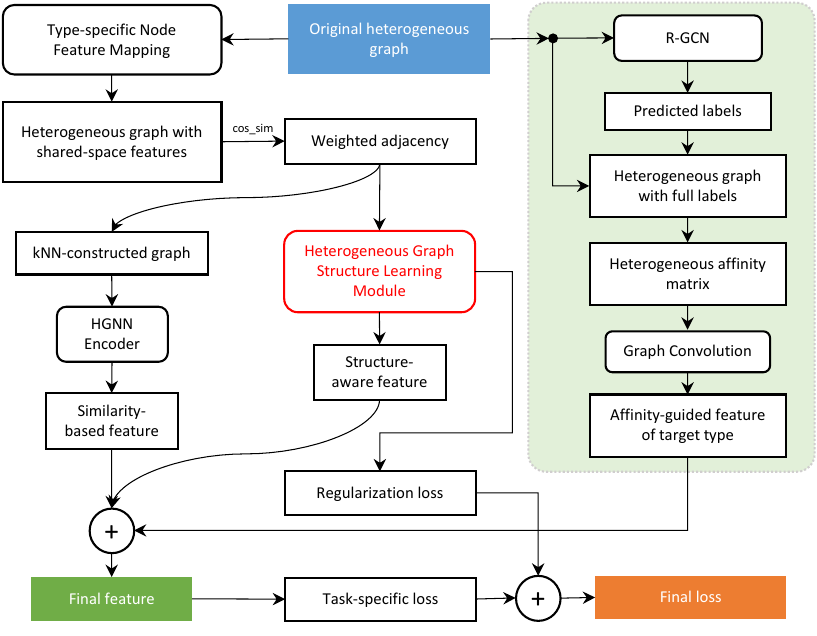}
    \caption{Overall learning framework of the proposed heterogeneous graph unified learning (HGUL). The green box area is the heterogeneous affinity learning module.}
    \label{fig:fw}
\end{figure}
The overall framework is designed to address the challenges of heterophily and structural noise in heterogeneous graphs by jointly learning structure-aware representations and refining relational dependencies. We refer to this framework as Heterogeneous Graph Unified Learning (HGUL), as it simultaneously tackles both heterophily and noisy connections in a unified manner. As illustrated in Figure~\ref{fig:fw}, the framework consists of three key stages: (1) similarity-based graph construction, (2) structure-aware representation learning, and (3) affinity-guided refinement with supervised optimization.
Given an original heterogeneous graph, nodes of different types are first projected into a shared feature space of dimension $d$ through a type-specific mapping function. This transformation ensures that node attributes from different types become comparable while preserving their semantic distinctions.
Formally, for each node type \(t\in\mathcal{A}\), we compute
\begin{equation}
\mathbf{h}_t\gets MLP(\mathbf{X}_t)\in\mathbb{R}^{n_t\times d}
\end{equation}
Based on the unified feature space, pairwise similarities between nodes are computed (e.g., via cosine similarity) under the guidance of the meta-relation set. This results in a weighted adjacency matrix that reflects cross-type semantic proximity rather than relying solely on the original graph topology.

To effectively balance robustness and structural expressiveness, the framework employs two complementary learning paths.
(1) Similarity-based Feature Learning:
When the original graph structure is heavily corrupted by noise, directly relying on it can significantly degrade model performance. To address this issue, we construct a $k$-nearest neighbor (kNN) graph based on the learned similarity matrix. Specifically, for each meta-relation $(i,j)$, nodes of type $i$ (or $j$) select their top-$k$ most similar neighbors from type $j$ (or $i$), forming a refined edge set.
This kNN-based construction captures local feature consistency and provides a robust alternative to unreliable edges. Moreover, in scenarios where original connections are sparse or missing, the kNN graph re-establishes meaningful interactions and enables effective feature aggregation across node types.
The resulting kNN graph emphasizes feature similarity while discarding noisy structural signals. We then apply a heterogeneous graph neural network (HGNN) encoder on this graph to obtain similarity-based representations \(\mathbf{h}_{kNN}\).
2) Structure-aware Graph Learning. 
Despite the presence of noise, the original graph still encodes valuable prior knowledge about relationships. To leverage this information, we feed the weighted adjacency matrix into a heterogeneous graph structure learning module. This module learns latent structural dependencies by adaptively refining the graph topology, producing structure-aware representations \(\mathbf{h}_{HGSL}\). 
Additionally, the module introduces a regularization term \(\mathcal{L}_{reg}\), 
The representations from the two paths are then combined:
\begin{equation}
\mathbf{h}=Norm(\mathbf{h}_{HGSL}+\mathbf{h}_{kNN})
\end{equation}

To explicitly address heterophily, we further incorporate label-aware information through an affinity-guided mechanism. This component leverages supervision signals to model relationships between nodes that may not share similar features but are semantically related.
The affinity-guided representation $\mathbf{h}_{aff}$ is fused with the learned representation via a gating mechanism:
\begin{equation}
\mathbf{h}_t=(1-g)\odot\mathbf{h}_t+ g\odot\mathbf{h}_{aff}
\end{equation}
where the gate g is defined as \(g=\sigma([\mathbf{h}_t;\mathbf{h}_{aff}]W+b)\)
with learnable parameters \(W\in\mathbb{R}^{2d\times d}\) and \(b\in\mathbb{R}^d\), and $\sigma(\cdot)$ denoting the sigmoid function. This mechanism adaptively balances structural features and label-informed affinities.

Then we obtain the task-specific loss using the final feature and calculate the weighted sum of task-specific loss and regularization loss to obtain the final loss:
\begin{equation}
    \mathcal{L}_{task}\gets\ell (\mathbf{h}_t, \mathbf{Y}_t)
\end{equation}
where $\mathbf{Y}_t$ denotes the ground-truth labels for target node type $t$.

The overall training objective combines the task loss with the structural regularization term:
\begin{equation}
    \mathcal{L}=\mathcal{L}_{task}+\gamma\mathcal{L}_{reg}
\end{equation}
where \(\gamma\) is a hyperparameter that controls the trade-off between predictive accuracy and structural robustness.
The entire framework is trained end-to-end, enabling joint optimization of feature transformation, structure learning, and affinity refinement. This unified design allows HGUL to effectively mitigate noise while capturing heterophilous relationships in heterogeneous graphs.

\subsection{Heterogeneous Graph Structure Learning}
To mitigate the adverse effects of noisy or unreliable connections in heterogeneous graphs, we employ a graph structure learning strategy that jointly optimizes the adjacency structure and node representations. This allows the model to adaptively refine the graph topology while learning expressive embeddings.
Given the weighted adjacency matrix produced by the preceding projection module, we treat each entry as a learnable parameter and denote the resulting matrix as \(A\).
This initialization provides a meaningful prior that reflects the similarity between node features while still allowing flexibility for subsequent optimization.

To suppress noisy edges, we perform stochastic sampling over the edge set.
A naive discrete sampling process, however, is non-differentiable and prevents gradient-based optimization of the graph structure. To address this issue, we adopt the Gumbel-Sigmoid reparameterization technique \citep{jang2017categorical} to approximate discrete sampling in a differentiable manner.
Specifically, for each edge \((u,v)\triangleq i\in\mathcal{E}\), we compute
\begin{equation}
    y_i=\sigma((\pi_i+g_i)/\tau)
\end{equation}
where $\pi_i=\sigma(A_{uv})$ and \(g_i\) is sampled from the Gumbel distribution, which can be generated as \(g_i=-\log(-\log\epsilon)\), \(\epsilon\sim U(0,1)\). \(\tau\) is the temperature scale parameter.
In practice, we employ the straight-through estimator (STE) \citep{jang2017categorical} to enable gradient backpropagation while maintaining discrete-like behavior in the forward pass.
Based on the sampled edge weights, we further apply a hard thresholding operation to filter out potentially noisy edges:
\begin{equation}
z_i=H(y_i-\delta)
\end{equation}
where \(H(\cdot)\) denotes the Heaviside step function and \(\delta\) is a predefined threshold.
It's worth noting that even though $\tau$ and $\delta$ jointly determine that whether to retain one edge, they can not be merged into one as $\tau$ also balances the model's exploration and convergence of graph structure by controlling the smoothness of gradients during the training process.
We then perform graph convolution over the refined adjacency matrix to obtain structure-aware node representations:
\begin{equation}
    h_{HGSL}=\text{R-GCN}(X,\tilde{A})
\end{equation}
Here, we use the original node features instead of the projected features to avoid potential gradient inconsistencies between edge weight learning and downstream representation learning.
To prevent the learned structure from deviating excessively from the initial graph prior, we introduce a regularization term:
\begin{equation}
    \mathcal{L}_{reg}=\|\tilde{A}-A\|^2
\end{equation}
The overall procedure is described in Algorithm~\ref{alg:hgul}.

\begin{algorithm}[t]
\LinesNumbered
\caption{Heterogeneous Graph Structure Learning (HGSL)}
\label{alg:hgul}
\KwIn{Node features $X$, initial weighted adjacency $A$ where $A_{ij} = \mathrm{cos\_sim}(h_i, h_j)$, temperature $\tau$, threshold $\delta$}
\KwOut{Structure-aware node representations $H_{HGSL}$}

\For{each edge $i \in \mathcal{E}$}{
    Compute edge probability $\pi_i$ from $A$\;
    
    $g_i = -\log(-\log(\epsilon))$, $\epsilon \sim U(0,1)$\; \tcp{Sample Gumbel noise}
    $y_i = \exp\left((\log \pi_i + g_i)/\tau\right)$
    
    $\tilde{y}_i = \mathbf{1}(y_i > \delta)$ \tcp{hard thresholding}
    
    $z_i = y_i+\tilde{y}_i-y_i$.detach() \tcp{straight-through estimator}
}

Construct binary edge mask $Z$ from $\tilde{y}_i$\;

\textbf{Graph Refinement:} Update adjacency matrix $\tilde{A} = A \odot Z$\;

$H_{HGSL} = \text{R-GCN}(X, \tilde{A})$ \tcp{Relational Graph Convolution}

\Return{$H_{HGSL}$}
\end{algorithm}

\subsection{Heterogeneous Affinity Learning}
Heterophily in heterogeneous graphs exhibits fundamentally different characteristics compared to homogeneous settings. Recent studies extend heterophily metrics to heterogeneous graphs by aggregating statistics over predefined meta-paths \citep{lin2025heterophily}, revealing that different relations may exhibit varying levels of heterophily. This observation highlights the need for a learning mechanism that can adaptively capture diverse relational patterns across node and edge types.
To address this, we propose a Heterogeneous Affinity Matrix, which provides a stronger structural prior for modeling class-level interactions. We elaborate our approach in two steps: (1) extending existing affinity definitions to incorporate higher-order structures, and (2) adapting the formulation to heterogeneous graphs.
A representative affinity definition is introduced in prior work \citep{zhu2021graph}, where class-level affinity is computed based on one-hop interactions. However, such formulations largely ignore higher-order structural information and tend to degenerate into degree-based statistics, which may lead to biased estimates of inter-class relationships.
To overcome this limitation, we incorporate multi-hop interactions into affinity modeling. At the same time, we enforce that the influence of longer paths decays with increasing hop distance. Formally, we consider a polynomial graph kernel:
$ker(A)=\{\sum_{k=0}^{\infty}\alpha_k A^k\}$, with $\alpha_k > \alpha_{k+1}$.
By choosing $\alpha_k = \alpha^k$ with $0 < \alpha < 1$, the formulation reduces to a Personalized PageRank (PPR) \citep{gasteiger2019predict,chien2021adaptive} kernel, which naturally balances local and global information.
Based on this construction, we define the following extended affinity matrix.
\begin{definition}[Extended affinity matrix] 
\begin{equation} \label{eq:cls_mat}
    \mathbf{C}_{cls}=\mathbf{Y}^\top ker(\mathbf{A})\mathbf{Y}=\mathbf{Y}^\top(\mathbf{I}-\alpha\mathbf{D}^{-1/2}\mathbf{A}\mathbf{D}^{-1/2})^{-1}\mathbf{Y}
\end{equation}
where \(0<\alpha<1\) is the damping factor, \(\mathbf{D}\in\mathbb{R}^{n\times n}\) is the diagonal node degree matrix, and \(\mathbf{Y}\in\{0,1\}^{n\times C}\) is the label indicator matrix. 
\end{definition}
Each entry \(\mathbf{C}_{cls}(i,j)\) quantifies the affinity between class $i$ and class $j$, aggregating contributions from paths of all lengths with exponentially decaying weights. This formulation naturally captures both homophilic and heterophilic interactions, while allowing different class pairs to exhibit distinct structural relationships.

In semi-supervised settings, the full label matrix Y is unavailable, making direct computation of \(\mathbf{C}_{cls}\) infeasible. To address this, we estimate the affinity matrix via a pretraining stage.
Specifically, we first train a GNN on the original heterogeneous graph to obtain node embeddings \(\mathbf{E}\in\mathbb{R}^{n\times d}\). A readout function followed by a softmax layer produces class probability estimates:
\begin{equation}
    \hat{\mathbf{Y}}=\textsc{Softmax}(\textsc{Readout}(\mathbf{E}))
\end{equation}

Without loss of generality, we reorder the nodes and partition the predictions into labeled and unlabeled subsets
\(\hat{\mathbf{Y}}=\begin{bmatrix}
    \hat{\mathbf{Y}}_{train} \\
    \hat{\mathbf{Y}}_{unknown}
\end{bmatrix}\). 
After convergence, we construct a surrogate label matrix by combining ground-truth labels and predictions: \(\begin{bmatrix}
    \mathbf{Y}_{train} \\
    \hat{\mathbf{Y}}_{unknown}
\end{bmatrix}\). Substituting it into the affinity formulation (Equation~\ref{eq:cls_mat}) yields an estimated matrix \(\hat{\mathbf{C}}_{cls}\), which approximates the true class-level affinity. 
We then use this matrix to generate affinity-guided features for the target node type through graph convolution:
\begin{equation}
    \mathbf{H}_{aff}\gets \hat{\mathbf{C}}_{cls}\mathbf{X}_t^{(l-1)}
\end{equation}

Then we propose to extend the proposed extended affinity matrix to heterogeneous graphs.
First, we replace the pretraining backbone with an R-GCN to capture relation-specific patterns. However, two key challenges arise:
(1) node labels are typically defined only for a specific target node type, while other node types serve as auxiliary context;
(2) a simple binary adjacency matrix is insufficient to encode heterogeneous semantics.
To address these issues, we construct a relation-aware augmented adjacency matrix. Specifically, we introduce a learnable matrix $R\in\mathbb{R}^{|\mathcal{A}|\times|\mathcal{A}|}$, where each entry \(R_{ij}\) represents the importance of interactions between node types i and j. The adjacency matrix is then reweighted as: $\bar{A}_{ij}\gets A_{ij} R_{\phi(i),\phi(j)}$
Next, we compute a PPR-based kernel over the reweighted adjacency $B=ker(\bar{A})$ which captures multi-hop interactions across heterogeneous relations.
Since supervision is only available for the target node type, we extract the corresponding submatrix \(B_t\in\mathbb{R}^{n_t\times n_t}\), where \(n_t\) is the number of target-type nodes.
Using this matrix and the label matrix $\mathbf{Y}_t$, we construct the heterogeneous affinity matrix:
\begin{equation}
\hat{\mathbf{C}}_{cls}=\mathbf{Y}_t^\top B_t \mathbf{Y}_t
\end{equation}
This formulation captures how different classes of target nodes interact through heterogeneous multi-hop paths, effectively summarizing the influence of auxiliary node and relation types.

\subsection{Model Analysis}

\paragraph{Adapt to inductive learning}
For large-scale heterogeneous graphs, inductive learning is essential for scalable training. Common sampling strategies include random-walk-based samplers \citep{lovasz1993random}, neighborhood sampling, and HGT sampling \citep{hu2020heterogeneous}. These methods typically start from a mini-batch of target-type nodes and iteratively expand neighborhoods through different meta-relations to construct a sampled heterogeneous subgraph.
To preserve graph semantics while maintaining efficiency, a relatively small batch size with sufficiently large sampling width and depth can be adopted, allowing each sampled subgraph to capture rich multi-hop heterogeneous context.
HGUL naturally adapts to this setting. Specifically, both the kNN construction and HGSL modules operate on sampled subgraphs rather than the full graph. Moreover, instead of constructing a global affinity matrix, HGUL computes a subgraph-level heterogeneous affinity matrix, which models pairwise affinities only among target-type nodes in the current batch. This makes affinity-guided learning compatible with mini-batch training and scalable inductive inference.

\paragraph{Computational overhead analysis}
We analyze the computational cost of each component in HGUL.
The pretraining backbone (i.e., R-GCN) has per-epoch complexity $O(|\mathcal{E}|d)$.
Assume a sampled batch contains $n$ nodes and $m$ edges.
For the kNN construction branch, using brute-force similarity computation, the complexity across meta-relations is $d\sum_{(i,j)\in\mathcal{R}}n_in_j$, where $n_i$ denotes the number of sampled nodes of type $i$. 
In practice, this cost is manageable under mini-batch sampling. For very large graphs, approximate nearest neighbor methods such as HNSW \citep{malkov2018efficient} can further reduce the cost close to subquadratic or near-linear complexity.
The HGSL module requires \(O(md)\), which scales linearly with the number of sampled edges.
In the heterogeneous affinity learning module, to compute the graph kernel, we use power iteration to approximate the PPR kernel: $O(n\cdot k \cdot m)$, where $k$ is the maximum number of iterations. 
Constructing the class-level affinity matrix requires
$O(n_0^2C)$, where $n_0$ is the number of target-type nodes in the original batch.
Thus, the affinity learning branch has complexity $O(n\cdot k \cdot m+n_0^2C)$.
Combining all components, the complexity of HGUL is: \(O(d(m+\sum_{(i,j)\in\mathcal{R}}n_in_j+n_0^2)+nkm+n_0^2C)\).
Since in practice $k$ and $C$ are small constants, and mini-batch sampling keeps $n$ and $m$ bounded, the dominant cost is near-linear in the sampled subgraph size. This ensures the scalability of HGUL to large heterogeneous graphs.

\section{Experiments}




\subsection{Experiment Settings}
\paragraph{Datasets}
For the datasets, we adopt a subset of the H2GB benchmark \citep{lin2025heterophily}, which contains large-scale graph learning dataset suite designed to evaluate models on graphs that exhibit both heterogeneity and heterophily. It consists of multiple datasets spanning diverse domains such as finance, cybersecurity, academic networks, e-commerce, and social networks. The evaluation metric for RCDD, IEEE-CIS-G and PDNS is F1 score due to class imbalance, and the others are accuracy. Due to the relatively large size of these datasets, we employ the HGT sampler \citep{hu2020heterogeneous} to perform recursive relation-aware neighbor sampling and construct subgraphs for mini-batch training.

\paragraph{Baselines} 
We compare our method against several representative and strong heterogeneous graph learning models. These include classical HGNNs including relational GNNs \citep{schlichtkrull2018modeling}, HAN \citep{wang2019heterogeneous}, HGT \citep{hu2020heterogeneous}, Simple-HGN \citep{platonov2023critical} and heterogeneous graph learning models that explicitly consider heterophily: HDHGR, LatGRL and Hetero2Net and H2G-former.
We also include typical homogeneous graph learning models: GCN \citep{kipf2017semi}, GraphSAGE \citep{hamilton2017inductive}, GAT \citep{velivckovic2018graph}, GIN \citep{xu2019powerful}, APPNP \citep{gasteiger2019predict}, NAGphormer \citep{chen2023nagphormer}, MixHop \citep{abu2019mixhop} and LINKX \citep{lim2021large} as comparison.
To evaluate robustness, we additionally compare with several representative robust graph learning and graph structure learning methods, including DropEdge \citep{rong2020dropedge}, LDS \citep{franceschi2019learning}, NeuralSparse \citep{zheng2020robust}, Pro-GNN \citep{jin2020graph}, PTDNet \citep{luo2021learning}.
We also include HGSL \citep{zhao2021heterogeneous}, a heterogeneous graph structure learning method, and PT-HGNN \citep{jiang2021pre}, which performs heterogeneous graph sparsification.

\paragraph{Experimental details}
We use Pytorch to implement the proposed method and conduct all experiments.  Experiments are conducted on Ubuntu 22.04.4 LTS with a single NVIDIA RTX 6000 Ada (48GB) and 256GB RAM. For all HGNN models, we choose the number of message passing layers in \{1, 2\} to avoid over-smoothing, the number of hidden dimensions in \{32, 64, 128\} and the number of attention heads in \{1, 2, 4, 8\} if there is attention mechanism.
Additionally, we apply a GraphNorm layer \citep{cai2021graphnorm} to the output embedding to improve numerical stability.
We use Adam optimizer \citep{kingma2014adam} with learning rate and weight decay searched in \{4e-3, 5e-3, 6e-3\} and \{0.0, 0.001\} respectively. In order to mitigate the impact of randomness, we select seed from \{0, 1, 42\} and search the optimal parameter space. Then we repeat the experiment 10 times to report the average results.
We adopt annealing schedule for the temperature hyperparameter \(\tau\) and set the damping factor \(\alpha=0.85\).

\subsection{Main Results}

\begin{table*}[t]
\caption{Results on heterogeneous graphs with different homophily ratio. OOM means out-of-memory. The best and second best results are \textbf{bolded} and {\ul underlined}.}
\label{tab:h2}
\begin{tabular}{cccccccc}
\toprule
Model & ogbn-mag & mag-year & oag-cs & RCDD & IEEE-CIS-G & H-Pokec & PDNS \\ \midrule
\multicolumn{8}{c}{\cellcolor[HTML]{D0D0D0}homogeneous graph   learning methods} \\
GCN & 42.90 ± 0.50 & 32.91 ± 0.50 & 18.22 ± 0.60 & 70.63 ± 0.36 & 85.81 ± 0.87 & 28.79 ± 1.07 & 81.22 ± 0.30 \\
GraphSAGE & 40.80 ± 0.56 & 36.28 ± 0.19 & 22.92 ± 0.29 & 77.29 ± 0.30 & 85.02 ± 0.83 & 31.49 ± 1.23 & 91.44 ± 0.32 \\
GAT & 48.60 ± 0.29 & 33.50 ± 0.62 & 19.12 ± 0.25 & 70.89 ± 0.20 & 86.71 ± 1.27 & 28.51 ± 0.45 & 93.97 ± 0.27 \\
GIN & 37.32 ± 0.33 & 31.15 ± 0.54 & 16.33 ± 1.34 & 74.72 ± 0.32 & 84.22 ± 0.34 & 28.53 ± 0.54 & 87.91 ± 0.46 \\
APPNP & 37.64 ± 0.31 & 29.79 ± 0.61 & 17.90 ± 0.60 & 57.27 ± 1.22 & 82.95 ± 0.67 & 27.27 ± 1.47 & 80.70 ± 0.73 \\
NAGphormer & 42.47 ± 0.74 & 32.60 ± 0.06 & 16.49 ± 0.55 & 80.59 ± 0.15 & 85.46 ± 0.50 & 17.07 ± 0.34 & 92.37 ± 0.22 \\
\midrule \multicolumn{8}{c}{\textit{heterophily   considered}} \\
MixHop & 46.99 ± 0.41 & 36.36 ± 0.28 & 23.04 ± 0.24 & 78.78 ± 0.27 & 85.43 ± 1.22 & 30.13 ± 0.86 & 92.78 ± 0.18 \\
LINKX & 40.83 ± 0.18 & 42.81 ± 0.14 & 15.32 ± 0.08 & 79.66 ± 0.94 & OOM & 31.42 ± 1.20 & 87.74 ± 0.52 \\
\midrule \multicolumn{8}{c}{\cellcolor[HTML]{D0D0D0}heterogeneous graph   learning methods} \\
R-GCN & 46.93 ± 0.46 & 35.60 ± 0.48 & 23.10 ± 1.09 & 78.05 ± 0.28 & {\ul 87.00 ± 1.35} & 31.44 ± 0.96 & 92.55 ± 0.44 \\
R-GraphSAGE & 50.94 ± 0.44 & 38.07 ± 0.41 & 22.81 ± 0.63 & 77.00 ± 0.32 & 86.81 ± 1.74 & 29.85 ± 0.47 & 92.81 ± 0.37 \\
R-GAT & 41.51 ± 0.47 & 35.40 ± 0.88 & 21.03 ± 0.59 & 67.17 ± 0.24 & 80.37 ± 0.62 & 22.09 ± 0.94 & 94.29 ± 0.16 \\
HAN & 39.00 ± 0.22 & 29.66 ± 0.43 & 13.14 ± 1.96 & 54.04 ± 2.17 & 78.56 ± 1.42 & 23.15 ± 0.43 & 84.58 ± 0.76 \\
HGT & 50.23 ± 0.48 & 39.47 ± 1.66 & 22.51 ± 0.40 & 78.91 ± 0.43 & 86.05 ± 1.01 & 30.89 ± 0.80 & 92.76 ± 0.15 \\
Simple-HGN & 43.39 ± 0.28 & 34.43 ± 1.23 & 22.03 ± 0.46 & 50.50 ± 0.89 & 79.67 ± 2.53 & {\ul 31.66 ± 0.86} & 89.33 ± 0.21 \\
\midrule \multicolumn{8}{c}{\textit{heterophily   considered}} \\
HDHGR & 49.21 ± 0.38 & 38.50 ± 0.30 & 22.12 ± 0.40 & 80.44 ± 0.25 & 84.79 ± 0.42 & 28.67 ± 0.19 & 93.96 ± 0.28 \\
LatGRL & 46.07 ± 0.40 & 35.19 ± 0.28 & 21.55 ± 0.36 & 80.01 ± 0.45 & 82.57 ± 0.64 & 29.09 ± 0.33 & 94.50 ± 0.53 \\
Hetero2Net & 51.60 ± 0.33 & 43.87 ± 0.25 & 24.99 ± 0.31 & {\ul 82.97 ± 0.37} & 85.06 ± 0.30 & 31.14 ± 0.59 & 95.72 ± 0.32 \\
H2G-former & {\ul 53.12 ± 0.32} & {\ul 45.28 ± 0.22} & {\ul 25.37 ± 0.28} & 81.03 ± 0.51 & 86.52 ± 0.34 & 31.01 ± 0.26 & \textbf{96.81 ± 0.09} \\
HGUL & \textbf{54.21 ± 0.27} & \textbf{46.11 ± 0.20} & \textbf{26.80 ± 0.22} & \textbf{85.10 ± 0.43} & \textbf{89.92 ± 0.48} & \textbf{33.05 ± 0.57} & {\ul 96.39 ± 0.23}
\\ \bottomrule
\end{tabular}
\end{table*}

We first evaluate the proposed HGUL framework on the original, noise-free datasets. The quantitative results are summarized in Table~\ref{tab:h2}. From the experimental results, we draw the following key observations.
1) HGUL outperforms all compared methods on the majority of datasets, demonstrating its effectiveness in learning robust representations for heterogeneous graphs with heterophily. The performance gain can be largely attributed to the proposed affinity-guided mechanism, which explicitly models class-level relationships and complements structural learning.
2) On datasets such as IEEE-CIS-G and RCDD, which exhibit strong heterophily due to fraud-related connections, HGUL achieves significantly larger improvements over baseline methods. This suggests that the proposed framework is particularly effective in scenarios where connected nodes exhibit distinct semantics. The combination of affinity modeling and structure refinement enables HGUL to capture long-range and cross-class dependencies that are often overlooked by conventional methods.
3) HGUL maintains competitive performance across datasets of varying sizes, ranging from small-scale graphs to large real-world networks. This indicates that the model generalizes well under different data regimes and remains scalable in inductive settings. The use of kNN-based graph construction and sparsity-controlled structure learning further contributes to its computational efficiency and stability.
4) Across all datasets, methods specifically designed for heterogeneous graphs with heterophily awareness consistently outperform those that fall into the other three categories. This highlights two important insights:
Firstly, treating heterogeneous graphs as homogeneous leads to substantial information loss, as it ignores type-specific semantics and relation diversity.
Secondly, ignoring heterophily results in misleading neighborhood aggregation, where dissimilar nodes are incorrectly smoothed together.
These observations are consistent with prior findings in homogeneous graphs, but are further amplified in heterogeneous settings due to the additional complexity of node and relation types.

\begin{table*}[] 
\caption{Ablation study. The w/o GSL variant is constructed by replacing HGSL module with a R-GCN.}
\label{tab:h2-abl}
\centering
\begin{tabular}{cccccccc}
\toprule
 & ogbn-mag & mag-year & oag-cs & RCDD & IEEE-CIS-G & H-Pokec & PDNS \\
\midrule
HGUL & 54.21 ± 0.27 & 46.11 ± 0.20 & 26.80 ± 0.22 & 85.10 ± 0.43 & 89.92 ± 0.48 & 33.05 ± 0.57 & 96.39 ± 0.23 \\
w/o kNN & 54.11 ± 0.30 & 45.88 ± 0.19 & 26.75 ± 0.30 & 85.02 ± 0.28 & 89.87 ± 0.17 & 33.04 ± 0.14 & 96.21 ± 0.35 \\
w/o GSL & 53.98 ± 0.09 & 46.07 ± 0.37 & 26.73 ± 0.24 & 84.94 ± 0.19 & 89.80 ± 0.50 & 33.01 ± 0.15 & 96.36 ± 0.12 \\
w/o aff & 52.66 ± 0.20 & 44.29 ± 0.10 & 24.97 ± 0.38 & 81.55 ± 0.40 & 87.76 ± 0.22 & 31.79 ± 0.39 & 94.82 ± 0.26 \\
\midrule
\multicolumn{8}{c}{\textit{noise added   (perturbation rate=0.2)}} \\
HGUL & 53.50 ± 0.22 & 43.82 ± 0.31 & 24.67 ± 0.17 & 82.39 ± 0.59 & 87.12 ± 0.25 & 32.54 ± 0.28 & 94.93 ± 0.36 \\
w/o kNN & 52.72 ± 0.19 & 43.77 ± 0.42 & 23.90 ± 0.33 & 80.92 ± 0.23 & 86.54 ± 0.34 & 32.02 ± 0.16 & 94.91 ± 0.07 \\
w/o GSL & 51.78 ± 0.35 & 43.29 ± 0.24 & 23.08 ± 0.42 & 80.58 ± 0.21 & 86.77 ± 0.16 & 30.98 ± 0.37 & 94.05 ± 0.11 \\
w/o aff & 51.97 ± 0.15 & 43.82 ± 0.56 & 23.79 ± 0.38 & 80.99 ± 0.29 & 86.94 ± 0.13 & 31.03 ± 0.12 & 94.17 ± 0.19
\\ \bottomrule
\end{tabular}
\end{table*}
\begin{figure}[t]
    \centering
    \begin{subfigure}{0.48\linewidth}
        \centering
        \includegraphics[width=\linewidth]{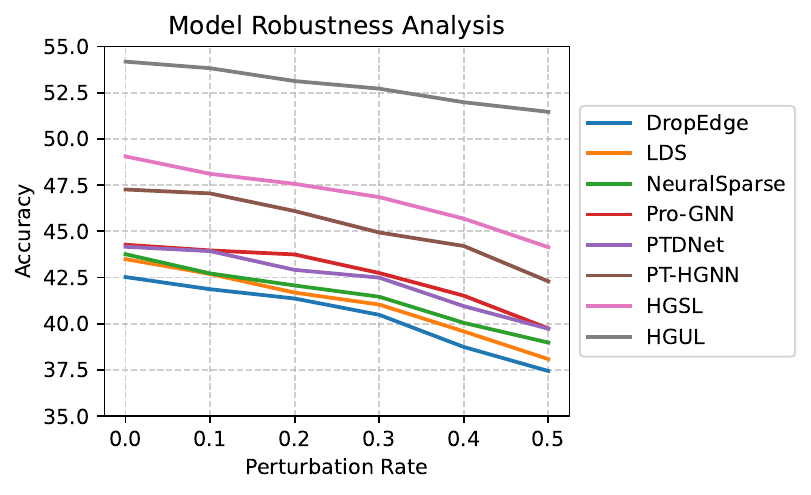}
        \caption{ogbn-mag}
    \end{subfigure}
    \begin{subfigure}{0.48\linewidth}
        \centering
        \includegraphics[width=\linewidth]{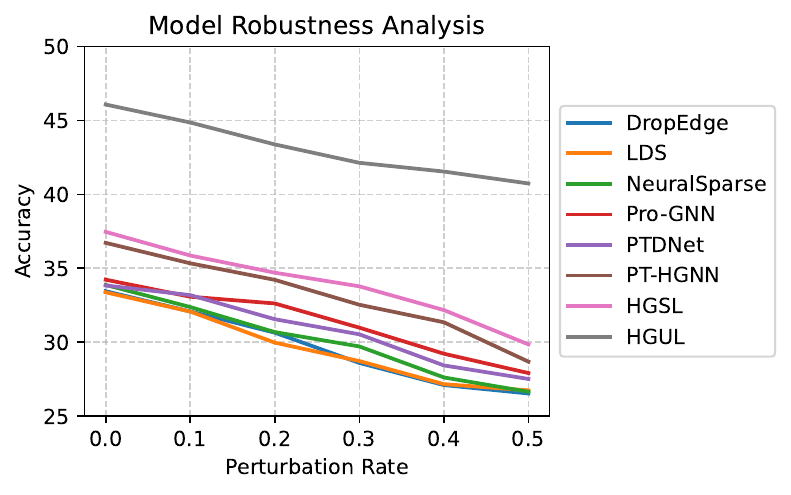}
        \caption{mag-year}
    \end{subfigure}
    \begin{subfigure}{0.48\linewidth}
        \centering
        \includegraphics[width=\linewidth]{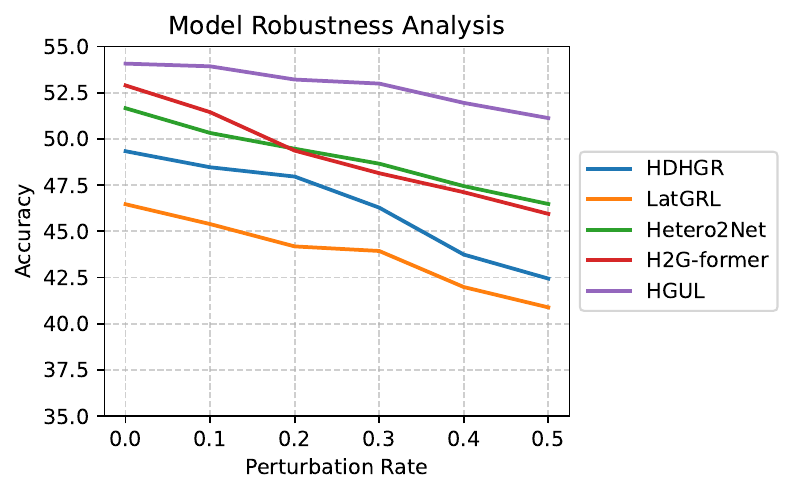}
        \caption{ogbn-mag}
    \end{subfigure}
    \begin{subfigure}{0.48\linewidth}
        \centering
        \includegraphics[width=\linewidth]{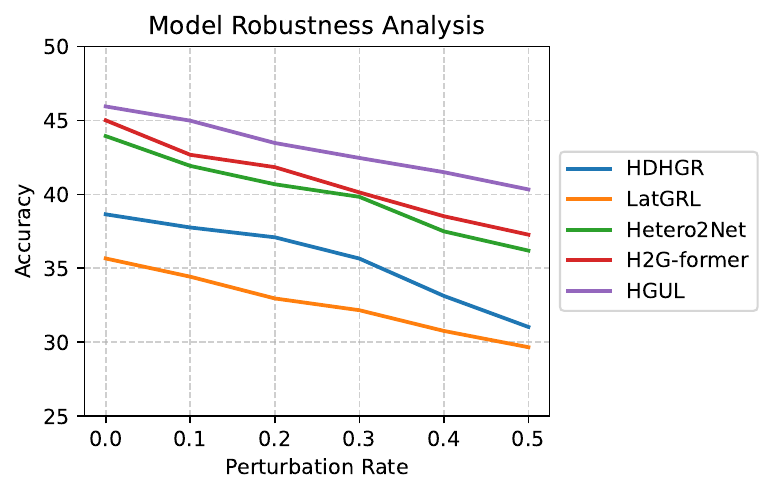}
        \caption{mag-year}
    \end{subfigure}
    \caption{Robustness of different learning methods against different level of noise. The upper row compares HGUL with competitive heterogeneous graphs learning methods, and the lower row compares HGUL with competitive graph structure learning methods.}
    \label{fig:robu}
\end{figure}
\paragraph{Robustness analysis under structural noise}
To further evaluate robustness, we inject synthetic noise into the graph structure by randomly perturbing edges, simulating adversarial or corrupted environments. The results are shown in Figure \ref{fig:robu}.
Overall, HGUL consistently outperforms basic robust learning methods and demonstrates significantly stronger robustness under structural perturbations. As the noise level increases, the performance of all methods degrades; however, HGUL exhibits a much slower decline, indicating its ability to effectively mitigate the impact of noisy connectivity.
In addition, consistent with previous observations, heterogeneous graph learning methods outperform homogeneous ones across all noise levels, highlighting the importance of explicitly modeling node and relation types.
Furthermore, by comparing HGSL with PT-HGNN, we observe clear advantages of GSL, which optimizes the graph topology under the supervision of downstream objectives and enables the model to preserve informative edges while suppressing misleading ones.
We also compare HGUL with other competitive methods. The results show that HGUL achieves the strongest robustness among all compared approaches, with the slowest performance degradation as the perturbation intensity increases. 

Taken together, the results validate the necessity of jointly modeling heterogeneity, heterophily, and structural noise. By integrating similarity-based learning, structure-aware refinement, and affinity-guided enhancement into a unified framework, HGUL is able to effectively capture complex relational patterns and achieve superior performance across diverse datasets.

\subsection{Ablation Studies}
To better understand the contribution of each component in HGUL, we conduct a series of ablation experiments by removing the three core modules: the kNN construction module, the GSL module, and the affinity-guided learning module. We evaluate these variants on both clean graphs and noisy graphs with a moderate perturbation rate of 0.2. The results are reported in Table~\ref{tab:h2-abl}.
Overall, performance consistently degrades when any component is removed, on both clean and noisy graphs, validating that all three modules contribute positively to the overall framework.
On clean graphs, the variant without affinity guidance (w/o aff) exhibits the most significant performance drop, and in several cases even falls below existing SOTA methods.
This observation highlights the critical role of affinity guidance in learning under heterophily.
Without the class-level prior provided by the estimated affinity matrix, the model struggles to distinguish cross-class interactions, resulting in inferior neighborhood aggregation.
This also suggests that when the original graph structure is relatively reliable, the primary performance gains come from affinity-guided message passing.
On noisy graphs, removing either the kNN module or the GSL module leads to substantially larger performance degradation compared to the clean setting. This confirms the importance of these two components in handling misleading connectivity and improving robustness.
In summary, the joint integration of the three modules is essential to the superior performance of HGUL.

\begin{figure}[t]
    \centering
    \begin{subfigure}{0.48\linewidth}
        \centering
        \includegraphics[width=\linewidth]{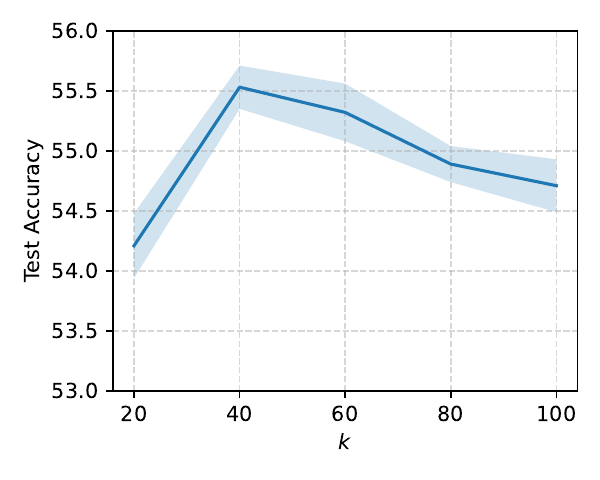}
        \caption{number of neighbors $k$}
    \end{subfigure}
    \begin{subfigure}{0.48\linewidth}
        \centering
        \includegraphics[width=\linewidth]{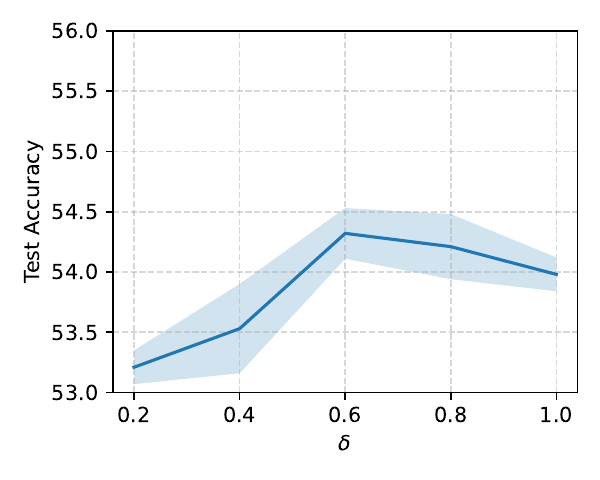}
        \caption{threshold value $\delta$}
    \end{subfigure}
    \begin{subfigure}{0.48\linewidth}
        \centering
        \includegraphics[width=\linewidth]{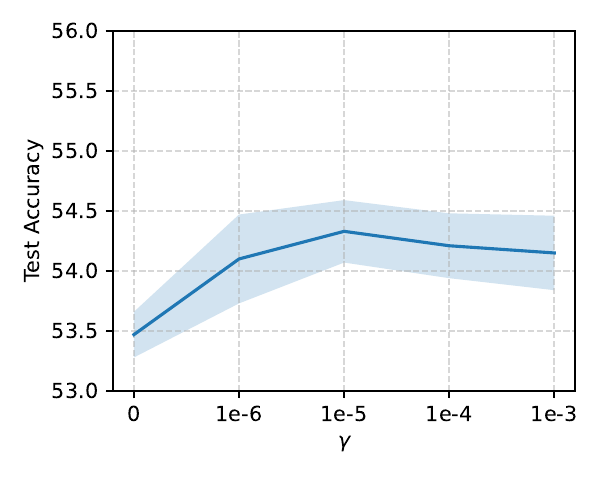}
        \caption{regularization coefficient $\gamma$}
    \end{subfigure}
    \caption{Parameter sensitivity analysis.}
    \label{fig:param}
\end{figure}
\subsection{Parameter Sensitivity Analyses}
We study the impact of three key hyperparameters in the proposed HGUL: \(k\) in kNN construction, the threshold \(\delta\) in the HGSL module and the coefficient of the regularization loss $\gamma$.
From Figure \ref{fig:param}, we observe that HGUL exhibits a stable performance trend across a wide range of parameter values, indicating that the framework is not overly sensitive to hyperparameter tuning. Nevertheless, each parameter plays a distinct role in controlling the balance between information preservation and noise suppression.
The parameter $k$ controls graph density by defining the number of neighbors per node. When $k$ is too small, the resulting graph is overly sparse, which restricts information propagation and leads to insufficient feature aggregation. As $k$ increases, the model benefits from richer neighborhood information, leading to improved performance. However, when $k$ becomes excessively large, the graph becomes overly dense, introducing noisy or less relevant connections.
The edge pruning threshold \(\delta\) in the HGSL module controls the sparsification of the learned graph by filtering edges based on their importance. The optimal performance is achieved at a moderate threshold, where noisy edges are effectively pruned while informative connections are preserved.
The regularization loss coefficient controls the strength of structural constraints imposed during training. When this coefficient is very small, the regularization term has little influence, and the model primarily optimizes the task-specific loss. In this case, the learned structure might deviate largely from the original. However, when $\gamma$ is too large, the regularization term dominates the optimization process. This restricts the model’s ability to adaptively refine the graph and may lead to underfitting.

\subsection{Efficiency Analyses}
\begin{table}[] \scriptsize
\caption{Efficiency Analysis. Training time per epoch (s).}
\label{tab:h2-effi}
\centering
\begin{tabular}{cccccc}
\toprule
Model & ogbn-mag & RCDD & IEEE-CIS-G & H-Pokec & PDNS \\ \midrule
GCN & 12.4 & 24.4 & 1.9 & 9.8 & 14.1 \\
R-GCN & 22.9 & 45.2 & 3.4 & 17.8 & 25.7 \\
HDHGR & 52.1 & 83.0 & 7.5 & 42.1 & 74.8 \\
LatGRL & 38.4 & 75.8 & 5.6 & 30.2 & 70.4 \\
Hetero2Net & 65.3 & 89.6 & 9.2 & 55.4 & 75.6 \\
H2G-former & 88.5 & 114.4 & 13.5 & 72.8 & 98.2 \\
HGUL & 34.0 & 67.1 & 5.4 & 23.8 & 64.5
\\ \bottomrule
\end{tabular}
\end{table}
We further evaluate the computational efficiency of the proposed framework (Table \ref{tab:h2-effi}). Compared with existing baseline methods, HGUL achieves strong efficiency due to its lightweight architecture. Specifically, both the kNN construction and HGSL module explicitly enforce sparsity, which reduces computational cost during message passing.
In addition, the affinity estimation is performed in an ultralight pretraining stage and reused during the main training process, which reduces repeated expensive computations.
Overall, HGUL achieves a favorable balance between expressiveness and efficiency.

\section{Conclusion}
In this paper, we study robust representation learning for heterogeneous graphs with heterophily, where noisy connectivity and cross-class interactions pose significant challenges to existing methods. To address this problem, we propose HGUL, a unified framework that jointly models similarity-based structure, adaptive graph refinement, and class-level affinity. By integrating kNN-based graph construction, graph structure learning, and heterogeneous affinity learning, HGUL effectively mitigates the impact of noisy edges while capturing heterophilous relationships.
Extensive experiments demonstrate that HGUL consistently improves performance on diverse graph benchmarks and remains robust under varying levels of structural noise. These results highlight the importance of jointly handling heterogeneity, heterophily, and noise in graph learning.

\small
\bibliographystyle{elsarticle-harv} 
\bibliography{main}

\appendix
\section{Spectral Investigation of Noise and Heterophily in Heterogeneous Graphs}
Previous works have discussed noisy connectivity~\citep{spielman2008graph,sadhanala2016graph,liu2021bridging,yu2022principle} and heterophily~\citep{bo2021beyond,luan2022revisiting} in homogeneous graphs from a spectral perspective. Building on a simple heterogeneous graph learning framework~\citep{yang2021interpretable}, we extend this line of analysis to HGNNs by examining how noise and heterophily manifest spectrally in the heterogeneous setting.
To simplify the discussion, we restrict attention to the case of two node types.
Let \(\mathbf{A}_1\in\mathbb{R}^{n_1\times n_1}\) and \(\mathbf{A}_2\in\mathbb{R}^{n_2\times n_2}\) denote the intra-type adjacency matrices for each node type, and \(\mathbf{B}\in\mathbb{R}^{n_1\times n_2}\) denote the inter-type adjacency matrix.
The full adjacency becomes:
\begin{equation}
\text{Adj}=
\begin{bmatrix}
    \mathbf{A}_1 & \mathbf{B} \\
    \mathbf{B}^\top & \mathbf{A}_2 \\
\end{bmatrix},
\end{equation}
and after applying any symmetric degree-based or attention-based normalization, we obtain:
\begin{equation}
\widehat{\text{Adj}}=
\begin{bmatrix}
    \hat{\mathbf{A}}_1 & \hat{\mathbf{B}} \\
    \hat{\mathbf{B}}^\top & \hat{\mathbf{A}}_2 \\
\end{bmatrix}.
\end{equation}
The corresponding (normalized) Laplacian is:
\begin{equation}
    \mathbf{L}=\mathbf{I}_{n_1+n_2}-\widehat{\text{Adj}}=
\begin{bmatrix}
    \mathbf{I}_{n_1}-\hat{\mathbf{A}}_1 & -\hat{\mathbf{B}} \\
    -\hat{\mathbf{B}}^\top & \mathbf{I}_{n_2}-\hat{\mathbf{A}}_2 \\
\end{bmatrix}
\end{equation}
This matrix admits a closed-form block diagonalization due to its symmetric two-block structure. Leveraging Schur complement, we obtain the decomposition of Laplace:
\begin{equation} \label{eq:diag}
\mathbf{L}=
\begin{bmatrix}
\mathbf{I}_{n_1} & \mathbf{0} \\
-\hat{\mathbf{B}}^\top(\mathbf{I}-\hat{\mathbf{A}}_1)^{-1} & \mathbf{I}_{n_2} \\
\end{bmatrix}
\begin{bmatrix}
\mathbf{I}_{n_1}-\hat{\mathbf{A}}_1 & \mathbf{0} \\
\mathbf{0} & \mathbf{S} \\
\end{bmatrix}
\begin{bmatrix}
\mathbf{I}_{n_1} & -(\mathbf{I}-\hat{\mathbf{A}}_1)^{-1}\hat{\mathbf{B}} \\
\mathbf{0} & \mathbf{I}_{n_2} \\
\end{bmatrix}
\end{equation}
where \(\mathbf{S}=(\mathbf{I}-\hat{\mathbf{A}}_2)-\hat{\mathbf{B}}^\top(\mathbf{I}-\hat{\mathbf{A}}_1)^{-1}\hat{\mathbf{B}}\).
From this decomposition, observe that the spectral properties of $\mathbf{L}$ are governed by two decoupled operators: the intra-type Laplacian $\mathbf{L}_1 \triangleq \mathbf{I}_{n_1} - \hat{\mathbf{A}}_1$ and the Schur complement $\mathbf{S}$. Let $\mathbf{L}_1 = \mathbf{U}_1 \boldsymbol{\Lambda}_1 \mathbf{U}_1^\top$ be its eigendecomposition with eigenvalues $0 \leq \lambda_1^{(1)} \leq \cdots \leq \lambda_{n_1}^{(1)}$. The correction term in $\mathbf{S}$ can then be expanded spectrally as:
\begin{equation}
\hat{\mathbf{B}}^\top \mathbf{L}_1^{-1} \hat{\mathbf{B}}
= \hat{\mathbf{B}}^\top \mathbf{U}_1 \boldsymbol{\Lambda}_1^{-1} \mathbf{U}_1^\top \hat{\mathbf{B}}
= \sum_{k=1}^{n_1} \frac{1}{\lambda_k^{(1)}} \left(\hat{\mathbf{B}}^\top \mathbf{u}_k\right)\left(\hat{\mathbf{B}}^\top \mathbf{u}_k\right)^\top,
\end{equation}
where $\mathbf{u}_k$ is the $k$-th eigenvector of $\mathbf{L}_1$. This reveals a frequency-dependent coupling: the cross-type connectivity $\hat{\mathbf{B}}$, projected onto the eigenbasis of the intra-type Laplacian, is weighted by the inverse eigenvalue $1/\lambda_k^{(1)}$—amplifying low-frequency modes and suppressing high-frequency ones.

\paragraph{Spectral signature of noise}
Suppose the observed adjacency is a noisy version of the ground truth: $\mathrm{Adj} = \mathrm{Adj}^* + \boldsymbol{\Delta}$, where $\boldsymbol{\Delta}$ is a symmetric random perturbation. After normalization, the Laplacian becomes $\mathbf{L} = \mathbf{L}^* + \delta\mathbf{L}$, and by Weyl's inequality,
\begin{equation}
|\lambda_k(\mathbf{L}) - \lambda_k(\mathbf{L}^*)| \leq \|\delta\mathbf{L}\|_2, \quad \forall\, k.
\end{equation}
Since random noise is spectrally unstructured—its energy distributes roughly uniformly across all eigenmodes—$\delta\mathbf{L}$ perturbs eigenvalues across the entire spectrum. In the block decomposition, noise contaminates all three sub-matrices $\hat{\mathbf{A}}_1, \hat{\mathbf{A}}_2, \hat{\mathbf{B}}$ simultaneously, shifting eigenvalues of both $\mathbf{L}_1$ and $\mathbf{S}$ at every frequency. Thus, \textit{noise manifests as a broadband perturbation across all spectral components}.

\paragraph{Spectral signature of heterophily}
Consider a graph signal $\mathbf{f} = [\mathbf{f}_1;\, \mathbf{f}_2]$ encoding node labels or features. The cross-type Dirichlet energy is:
$$
E_{\mathrm{cross}} = \sum_{(i,j)\in\mathcal{E}_{\mathrm{cross}}} \hat{B}_{ij}\, (f_{1,i} - f_{2,j})^2.
$$
Under homophily, cross-type edges connect similar nodes, so $E_{\mathrm{cross}}$ is small and the cross-type signal is smooth—its energy concentrates in low-frequency modes. Since $1/\lambda_k^{(1)}$ is large for small $\lambda_k$, the correction $\hat{\mathbf{B}}^\top \mathbf{L}_1^{-1}\hat{\mathbf{B}}$ is significant, and the Schur complement $\mathbf{S}$ reflects strong spectral coupling between the two node types.
Under heterophily, cross-type edges connect dissimilar nodes, so $E_{\mathrm{cross}}$ is large and the cross-type signal is non-smooth. The projection $\hat{\mathbf{B}}^\top \mathbf{u}_k$ concentrates on eigenvectors with large $\lambda_k^{(1)}$ (high frequencies), but these are precisely the modes suppressed by $1/\lambda_k^{(1)}$. Consequently, the correction term shrinks:
$$
\hat{\mathbf{B}}^\top \mathbf{L}_1^{-1} \hat{\mathbf{B}} \approx \sum_{k:\,\lambda_k^{(1)}\,\text{large}} \frac{1}{\lambda_k^{(1)}} \left(\hat{\mathbf{B}}^\top \mathbf{u}_k\right)\left(\hat{\mathbf{B}}^\top \mathbf{u}_k\right)^\top \ll \sum_{k} \frac{1}{\lambda_k^{(1)}} \left(\hat{\mathbf{B}}^\top \mathbf{u}_k\right)\left(\hat{\mathbf{B}}^\top \mathbf{u}_k\right)^\top,
$$
and the Schur complement approaches $\mathbf{S} \approx \mathbf{I}_{n_2} - \hat{\mathbf{A}}_2$, effectively decoupling the two node types. Meanwhile, message passing via $\widehat{\mathrm{Adj}}$ aggregates dissimilar cross-type features:
$$
\left(\widehat{\mathrm{Adj}}\,\mathbf{f}\right)_{\text{type-1}} = \hat{\mathbf{A}}_1 \mathbf{f}_1 + \hat{\mathbf{B}}\,\mathbf{f}_2,
$$
where the term $\hat{\mathbf{B}}\,\mathbf{f}_2$ injects high-frequency energy into the type-1 representation (since connected cross-type features are dissimilar). Thus, \textit{heterophily manifests as a structured perturbation concentrated in the high-frequency spectral components}, in contrast to the broadband signature of noise.

\begin{figure}[t]
\centering
\begin{subfigure}[]{0.48\linewidth}
    \centering
    \includegraphics[width=\linewidth]{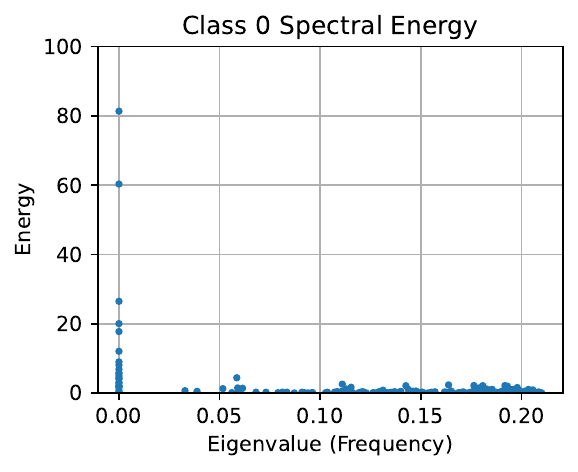}
    \caption{IMDB}
\end{subfigure}
\begin{subfigure}[]{0.48\linewidth}
    \centering
    \includegraphics[width=\linewidth]{figures/IMDB_spectral_energy.pdf}
    \caption{IMDB with noise}
\end{subfigure}
\begin{subfigure}[]{0.48\linewidth}
    \centering
    \includegraphics[width=\linewidth]{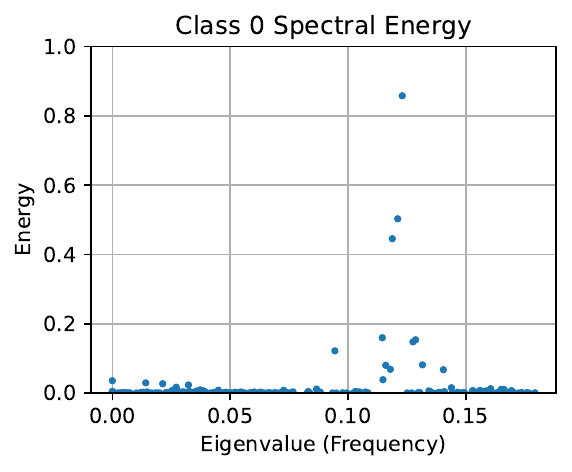}
    \caption{ogbn-mag}
\end{subfigure}
\begin{subfigure}[]{0.48\linewidth}
    \centering
    \includegraphics[width=\linewidth]{figures/obgn_spectral_energy.pdf}
    \caption{ogbn-mag with noise}
\end{subfigure}
\caption{Spectral signature of heterogeneous graphs. We adopt IMDB and obgn-mag as examples of homophily and heterophily dominated examples, respectively.}
\label{fig:spect}
\end{figure}
\paragraph{Visualization}
To corroborate the theoretical analysis, we conduct spectral analyses on two real-world heterogeneous graph benchmarks and visualize the distribution of graph signal energy across eigenvalues in Figure~\ref{fig:spect}. 
On the IMDB dataset, the spectral energy is sharply concentrated at the lowest eigenvalues, indicating that node labels vary smoothly across the graph. This is consistent with a strongly homophilic structure in which simple low-pass filters—such as standard GCN aggregation—are expected to perform well.
In contrast, the OGBN-MAG dataset exhibits a distinct energy peak around eigenvalue $\approx 0.12$, reflecting a non-trivial spectral mass at intermediate-to-high frequencies.
This shift is attributable to the greater relational diversity in OGBN-MAG (e.g., there are a large number of interdisciplinary citations in the paper-cite-paper relationship), whose heterogeneous semantics introduce heterophilic structure—high-frequency variations that naive smoothing-based aggregation would suppress.
These observations align with our theoretical finding that heterophily concentrates signal energy in high-frequency modes, whereas homophilic graphs are dominated by low-frequency components.
In addition, the impact of noise on both datasets is similar, making the energy more dispersed in the spectral domain.

Through the discussion in this section, it can be realized that the two factors---noise and heterophily---are fundamentally different in the spectral domain.
This confirms the rationality of the proposed model in handling two factors in separate modules.
\end{document}